\documentclass{article}

\usepackage{PRIMEarxiv}

\usepackage[utf8]{inputenc} % allow utf-8 input
\usepackage[T1]{fontenc}    % use 8-bit T1 fonts
\usepackage{hyperref}       % hyperlinks
\usepackage{url}            % simple URL typesetting
\usepackage{booktabs}       % professional-quality tables
\usepackage{amsfonts}       % blackboard math symbols
\usepackage{nicefrac}       % compact symbols for 1/2, etc.
\usepackage{microtype}      % microtypography
\usepackage{xcolor}         % colors

\usepackage{graphicx}
\usepackage{wrapfig}
\usepackage{subfig}

\usepackage{cite}

\usepackage{amsmath}
\usepackage{bm}
\usepackage{amssymb}

\usepackage{algorithm}
\usepackage{algorithmic}

\usepackage{array}
\usepackage{multirow}
\usepackage{makecell}
\usepackage{arydshln}
\title{Enhancement Encoding: A Novel Imbalanced Classification Approach via Encoding the Training Labels
}

\author{%
  Jia-Chen Zhao\\
  School of Information Science and Engineering\\
  Shandong University\\
  \texttt{jc\_zhao@mail.sdu.edu.cn}\\
}

\begin{document}
\maketitle

\begin{abstract}
  Class imbalance, which is also called long-tailed distribution, is a common problem in classification tasks based on machine learning. If it happens, the minority data will be overwhelmed by the majority, which presents quite a challenge for data science. To address the class imbalance problem, researchers have proposed lots of methods: some people make the data set balanced (SMOTE \cite{smote}), some others refine the loss function (Focal Loss \cite{focalLoss}), and even someone has noticed the value of labels influences class-imbalanced learning (Yang and Xu. \textit{Rethinking the value of labels for improving class-imbalanced learning.} In NeurIPS 2020 \cite{valueOfLabels}), but no one changes the way to encode the labels of data yet. Nowadays, the most prevailing technique to encode labels is the one-hot encoding due to its nice performance in the general situation. However, it is not a good choice for imbalanced data, because the classifier will treat majority and minority samples equally. In this paper, we innovatively propose the \textbf{enhancement encoding} technique, which is specially designed for the imbalanced classification. The enhancement encoding combines re-weighting and cost-sensitiveness, which can reflect the difference between hard and easy (or minority and majority) classes. To reduce the number of validation samples and the computation cost, we also replace the confusion matrix with a novel \textbf{soft-confusion matrix} which works better with a small validation set. In the experiments, we evaluate the enhancement encoding with three different types of loss. And the results show that enhancement encoding is very effective to improve the performance of the network trained with imbalanced data. Particularly, the performance on minority classes is much better.
\end{abstract}

% keywords can be removed
% \keywords{First keyword \and Second keyword \and More}

\section{Introduction}

\subsection{Class Imbalance Problem}

Today, it is widely accepted that the quantity and quality of training data are of great importance for machine learning (ML) \cite{imagenet, alexnet, vgg, googlenet}. A balanced data set helps a lot for ML, but in quite a few cases, researchers are obliged to process class-imbalanced samples \cite{LFID,fuzzySet}. Class imbalance describes the phenomenon that data of different classes is not uniformly distributed, i.e. instances in some classes are plentiful but others are lacking \cite{onTheClassImbalanceProblem, c45andImbalance}. When learning from a class-imbalanced training set, the general classifiers try to minimize loss or maximize accuracy on the imbalanced data. It may make the learner overfit the majority classes and underfit the minorities \cite{preliminaryStudy, theClassImbalanceProblem}. At inference time, the classifiers will prefer that an unknown instance belongs to the majorities \cite{balancingStrategiesAndClassOverlapping, c45andImbalance}, which declines the test accuracy of minority data.

\subsection{One-Hot Encoding}
\label{introduction_onehot}
One-hot encoding is one of the most prevailing techniques to encode labels or features now \cite{breastCancerClassification, autonomousNavigation, textEncoding, highCardinalityFeatures}. Compared with directly using the index of class as the label, the one-hot encoding technique expresses the label with a vector instead of a number. Suppose there is an instance belonging to class $p$. Its one-hot label can be described as:
\begin{gather}
    \bm{l} = [l_{0},\ l_{1},\ l_{2}, \cdots,\ l_{q}, \cdots,\ l_{N-1}]
,\qquad
    l_{q}= 
    \left\{
        \begin{array}{cc}
            0, & q \neq p\\
            1, & q = p\\
        \end{array}
    \right. 
\end{gather}
\noindent where $N$ is the total number of classes. Note that the indices of classes start at $0$ and end at $N\!-\!1$.

There are some significant advantages of one-hot encoding. Firstly, the label of any one class is ``equally'' different from all the other ones. More specifically, the Euclidean distance between any pair of different labels is always $\sqrt{2}$ and the Hemming distance is always $2$. Secondly, one-hot labels work very well with some other fashionable techniques such as \emph{Softmax} output activation, \emph{cross-entropy} loss, etc. That's why one-hot encoding is so popular in classification.

Nevertheless, one-hot encoding is not everything. There are still some problems: 

\begin{itemize}
\item \textbf{Sparsity:} The one-hot labels are too sparse. They are filled with plenty of ``0''s and a few ``1''s. Too much redundancy of information will eat up memory of the computer.

\item \textbf{Unfriendly to imbalanced data:} There are always some samples which are more difficult to be classified than others, but one-hot labels can not reflect this difference. This problem is particularly prominent when the data is class-imbalanced. 
\end{itemize}

So, a new encoding technique for labels is demanded.

\subsection{Our Work}

We have studied how to make full use of one-hot labels' redundancy to improve the performance of imbalanced classification, and in this paper, we innovatively propose the \textbf{enhancement encoding} technique. In contrast to some other methods for addressing the class imbalance problem (such as oversampling, re-weighting, etc.), it is not compulsory to learn about the distribution of training data for enhancement encoding. That means our method will be very effective when the data is not only imbalanced but also from a stream. 

In our algorithm, calculating the confusion matrix on the validation set is a necessary step. Meanwhile, enough validation instances per class are significant for an effective confusion matrix. Unfortunately, it conflicts with the lack of minority data. So, we propose the \textbf{soft-confusion matrix} to replace the confusion matrix. Soft-confusion matrix can work pretty well with a small validation set.

If you want to implement the enhancement encoding in your algorithm, you don't have to change the structure of your neural network or complicate the backpropagation (BP) too much. Re-encoding labels of the training set is all you need. Thanks to the soft-confusion matrix and fewer validation instances, a little bit of extra computation is just enough.

%%%%%%%%%%%%%%%%%%%%%%%%%%%%%%%%%%%%%%%%%%%%%%%%%%%%%%%%%%%%

\begin{wrapfigure}{0}
{0.5\linewidth}
\centering
\vspace{-1.7cm}
\includegraphics[width=1.\linewidth, trim = 80 70 48 50, clip ]{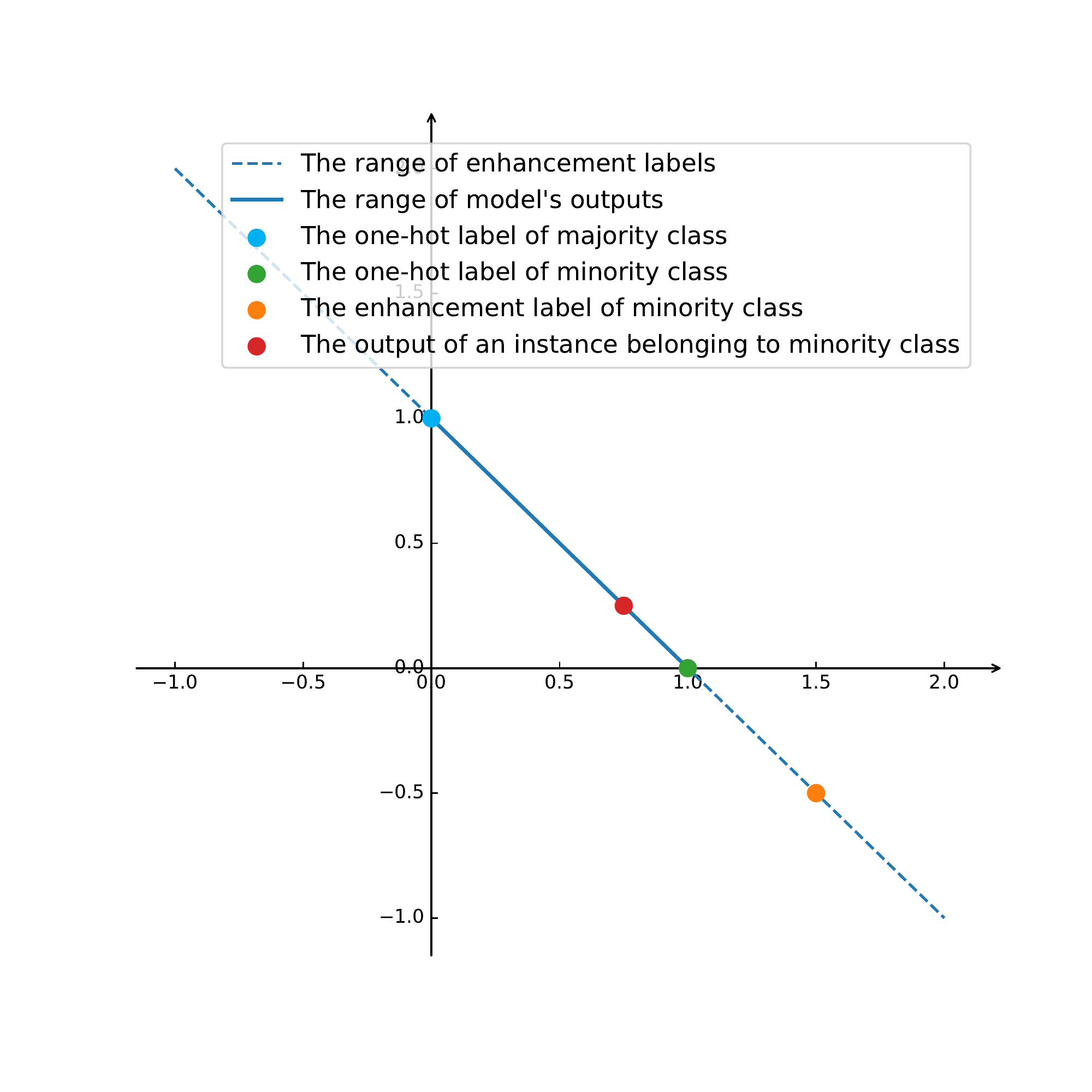} \caption{The basic idea of the enhancement label. With enhancement encoding, the label of minority class is moved from $(1,0)$ to the location on the dashed line.}
\label{BToEC}
\vspace{-1.2cm}
\end{wrapfigure}

\section{Proposed Approach}

\subsection{Soft-Confusion Matrix}
\label{sec_scm}

The confusion matrix is one of the most prevailing metrics in machine learning \cite{buildingText, rangeBasedConfusionMatrix, retinalImage, usingConfusionMatrices}. If there are $N$ classes in the classification, the confusion matrix denoted by $\bm{R}$ is an $N \times N$ matrix ($\bm{R} \in \mathbb{R}^{N \times N}$). Its element $r_{p,q}$ denotes the number of instances belonging to class $p$ but classified into class $q$. The greater the diagonal elements, the better the classifier performs.

The confusion matrix is pretty handy in classification, but there are still some shortcomings, such as working poorly when the data is insufficient. To improve the effect and make the most of data, we propose the soft-confusion matrix $\bm{S}$. $\bm{S}$ is the same size as $\bm{R}$ ($\bm{S} \in \mathbb{R}^{N \times N}$), and its elements can be described as:
\begin{equation} 
s_{p,q} = \sum_{i \in {\bf D}_p} P(q|{\bf x}_i)
\label{scme}
\end{equation}

\noindent where $P(q|{\bf x}_i)$ denotes the prediction probability that instance $i$ belongs to class $q$, ${\bf D}_p$ is a subset of the data set, which contains all instances belonging to class $p$. That is ``$i \in {\bf D}_p$'' means instance $i$ belongs to class $p$. An essential difference between the confusion matrix and the soft-confusion matrix is that the former counts the number of instances and the latter calculates the sum of probability. And obviously, the confusion matrix (without normalization) only contains integers, and the elements of the soft-confusion matrix can be non-integers. 

Certainly, the soft-confusion matrix can be calculated by nested loops, but there is a more efficient method via matrix multiplication:
\begin{equation} 
\bm{S} = {\bm L}^{\rm T} \bm{Y}
\label{scm}
\end{equation}
\noindent where $\bm{Y}$ contains outputs of the classifier and $\bm{L}$ contains one-hot encoded labels. More specifically, the element $y_{i,q}$ of $\bm{Y}$ is equal to $P(q|{\bf x}_i) $ and the element $l_{i,p}$ of $\bm{L}$ can be ``0'' (when instance $i$ does not belong to class $p$) or ``1'' (when instance $i$ belongs to class $p$). That is:

\begin{equation}
    y_{i,q}=P(q|{\bf x}_i)
,\qquad
    l_{i,p}= 
    \left\{
        \begin{aligned}
            &0, &&i \notin {\bf D}_p\\
            &1, &&i \in {\bf D}_p
        .\end{aligned}
    \right.
    \label{oh_label}
\end{equation}

The equivalence of Equation \eqref{scme} and Equation \eqref{scm} can be proved very easily:
\begin{gather}
    \bm{S} = {\bm L}^{\rm T} \bm{Y} \Longleftrightarrow s_{p,q} = \sum_{i}l_{i,p}\ y_{i,q}\ 
,\\
    % \begin{aligned}
    s_{p,q} 
    % &
    = \sum_{i}l_{i,p}\ y_{i,q}= \sum_{i}l_{i,p} P(q|{\bf x}_i) 
    % \\&
    =\sum_{i \in {\bf D}_p} P(q|{\bf x}_i)
    % .\end{aligned}
.\end{gather}
By the way, the method of Equation \eqref{scm} can also be used to calculate the general confusion matrix:
\begin{equation}
    y'_{i,q}= 
    \left\{
        \begin{aligned}
            &0, && q\neq\mathop{\arg\max}\limits_j\ y_{i,j} \\
            &1, && q=\mathop{\arg\max}\limits_j\ y_{i,j}\ 
        ,\end{aligned}
    \right.
\qquad
    \bm{R} = {\bm L}^{\rm T} {\bm Y}'
.\end{equation}

\subsection{The Basic Idea of Enhancement Encoding}

As discussed in Section \ref{introduction_onehot}, one-hot encoding is not a good choice for imbalanced learning. In an imbalanced classification task, identifying a minority instance is much more difficult than identifying a majority one, so it should be better to let the classifier pay more attention to learning from the minority data.

The enhancement encoding is an improved method adapted from the one-hot encoding. The graphical explanation for its basic idea is shown in Figure \ref{BToEC}. Because of circumscribed by the output layer's activation function (such as Softmax and Sigmoid), the elements of the output vector are all less than ``1'', greater than ``0'' and their sum is ``1'':
\begin{gather}
    0<y_{i,q} <1
,\qquad
    \sum_{q=0}^{N-1} y_{i,q}=1
.\end{gather} 
That means points representing outputs are always on the solid line in Figure \ref{BToEC}. Meanwhile, enhancement encoding moves the label of minority class from $(1,0)$ to the location on the dashed line. The process of training can be regarded as making the output and the label closer and closer together. The distance between them is related to the loss. So the new encoding method \emph{enhances} the part of loss caused by minority samples while training.

\begin{wrapfigure}{0}{0.62\linewidth}
\centering
\begin{minipage}{\linewidth}
\renewcommand{\algorithmicrequire}{\bf{Input:}}
\renewcommand{\algorithmicensure}{\bf{Output:}}
% \vspace{0.15cm}
\begin{algorithm}[H]
    \caption{\\ The BP algorithm with enhancement encoding}
    \label{the_algorithm}
    \begin{algorithmic}[1]
        \REQUIRE { \ \\
            ${\bm X}_{tr}$: Input values of the training set. \\
            ${\bm L}_{tr}$: One-hot encoded labels of the training set. \\
            ${\bm X}_v$: Input values of the validation set. \\
            ${\bm L}_v$: One-hot encoded labels of the validation set. \\
            $\eta$: Learning rate for the network's weights $\bm{\theta}$. \\
            $\mu$: Updating rate for the generator matrix $\bm G$. \\
            $\epsilon$: Enhancement rate. \\
            $E$: Number of training epochs. \\
            $b$: Batch size.
        }
        \ENSURE { \ \\
            $\bm{\theta}^*$: Weights of the network which have been trained.\\
            \ 
        }
        \STATE Construct the neural network.
        \STATE Initialize the network's weights $\bm{\theta}$.
        \STATE ${\bm G} \gets {\bm I}$ \hfill $\rhd$Initialize ${\bm G}$ as identity matrix
        \FOR {$i_{epoch}=1$ \TO $E$} 
            \STATE ${\bm Y}_v \gets$ forward\_pass\_without\_tracking(${\bm X}_v$, $\bm{\theta}$)
            \STATE ${\bm S} \gets$ soft\_confusion\_matrix(${\bm L}_v$, ${\bm Y}_v$) \hfill $\rhd$Eq. \eqref{scm}
            \STATE ${\bm S}' \gets$ normalize\_SCM(${\bm S}$) \hfill $\rhd$Eq. \eqref{normalize_SCM}
            \STATE ${\bm C} \gets$ cost\_matrix(${\bm S}'$) \hfill $\rhd$Eq. \eqref{cost_matrix}
            \STATE ${\bm G} \gets$ update\_generator\_matrix(${\bm G}$, ${\bm C}$, $\mu$, $\epsilon$) \hfill $\rhd$Eq. \eqref{update_generator_matrix}
            \STATE ${\bm L}^{\rm e}_{tr} \gets$ encode\_training\_labels(${\bm L}_{tr}$, ${\bm G}$) \hfill $\rhd$Eq. \eqref{encode_training_labels}
            \STATE $({\bf X}_{tr\rm B}, {\bf L}^{\rm e}_{tr\rm B}) \gets$ batch\_the\_training\_set(${\bm X}_{tr}$, ${\bm L}^{\rm e}_{tr}$, $b$)
            \FOR {$({\bm X}_{tr\rm b}, {\bm L}^{\rm e}_{tr\rm b})$ {\bf in} $({\bf X}_{tr\rm B}, {\bf L}^{\rm e}_{tr\rm B})$}
                \STATE ${\bm Y}_{tr\rm b} \gets$ forward\_pass(${\bm X}_{tr\rm b}$, $\bm{\theta}$)
                \STATE $J_{\rm b}^{\rm e} \gets$ loss(${\bm L}^{\rm e}_{tr\rm b}$, ${\bm Y}_{tr\rm b}$)
                \STATE $\nabla_{\bm \theta} J_{\rm b}^{\rm e} \gets$ backward\_pass($J_{\rm b}^{\rm e}$, $\bm \theta$, ${\bm X}_{tr\rm b}$) \hfill $\rhd$Gradient
                \STATE ${\bm \theta} \gets$ update\_the\_weights($\nabla_{\bm \theta} J_{\rm b}^{\rm e}$, ${\bm \theta}$, $\eta$)
            \ENDFOR 
        \ENDFOR 
        \STATE ${\bm \theta}^* \gets {\bm \theta}$
        \RETURN ${\bm \theta}^*$ 
    \end{algorithmic}
\end{algorithm}
\end{minipage}
\vspace{-0.7cm}
\end{wrapfigure}

\subsection{The Implementation of Enhancement Encoding}
\label{implementation}

In linear block code theory, the generator matrix $\bm{G}$ is usually utilized to encode the information:
\begin{equation} 
{\bm c} = {\bm m} {\bm G}
\label{encoding}
\end{equation}
\noindent where ${\bm m}$ is the message, and ${\bm c}$ is the channel code. Note that ${\bm m}$ and ${\bm c}$ in Equation \eqref{encoding} are row vectors. In this way, the labels can be encoded again to implement the enhancement encoding:
\begin{equation} 
{\bm l}^{\rm e}_i= {\bm l}_i \bm{G}
\label{encode_one_training_label}
\end{equation}
\noindent where ${\bm l}_i$ denotes the one-hot label of instance $i$ and ${\bm l}^{\rm e}_i$ denotes the enhancement label. It should be noted that the matrix multiplication in Equation \eqref{encoding} is defined over $\text{GF}(2)$, but multiplication in Equation \eqref{encode_one_training_label} is defined over the real number field. An advantage of the label generator matrix is the convenience to express enhancement encoding of multiple instances:
\begin{equation} 
{\bm L}^{\rm e}
\!=\!
\left[\!
    \begin{array}{c}
        {\bm l}^{\rm e}_0 \\
        {\bm l}^{\rm e}_1 \\
        {\bm l}^{\rm e}_2 \\
        \vdots \\ 
        {\bm l}^{\rm e}_i \\
        \vdots \\
    \end{array}
\!\right]
\!=\!
\left[\!
    \begin{array}{c}
        {\bm l}_0 \bm{G} \\
        {\bm l}_1 \bm{G} \\
        {\bm l}_2 \bm{G} \\
        \vdots \\ 
        {\bm l}_i \bm{G} \\
        \vdots \\
    \end{array}
\!\right]
\!=\!
\left[\!
    \begin{array}{c}
        {\bm l}_0 \\
        {\bm l}_1 \\
        {\bm l}_2 \\
        \vdots \\ 
        {\bm l}_i \\
        \vdots \\
    \end{array}
\!\right]\!\bm{G}
\!=\!
\bm{L}\!\bm{G}
.\end{equation}
However, how to get the generator matrix is still a question. There are some properties we wish the generator matrix to possess:

\begin{enumerate}
\item To re-weight the classes, the diagonal elements should be greater than ``1''.
\item To make the labels cost-sensitive, the non-diagonal elements should be less than ``0''.
\item To constrain the labels and the classifier outputs in the same hyperplane, as shown in Figure \ref{BToEC}, the sum of elements in the same row should be ``1''.
\end{enumerate}

That is:
\begin{gather}
    \quad g_{p,p} >1, \label{generator_matrix_property_1}
\\
    \quad g_{p,q} <0 \quad (p \neq q), \label{generator_matrix_property_2}
\\
    \sum_{q=0}^{N-1} g_{p,q} =1 
    .\label{generator_matrix_property_3}
\end{gather} 
In consideration of these conditions, we design an algorithm to calculate and update the generator matrix dynamically based on the soft-confusion matrix.

We add some extra steps before a training epoch. Firstly, compute the soft-confusion matrix of the network on the validation set and normalize it. The normalization can be described as follows: 
\begin{equation}
s_{p,q}' = \frac{s_{p,q}}{n_p}
\label{normalize_SCM}
\end{equation}

\noindent where $n_p$ is the number of instances belonging to class $p$ in the validation set. The normalization is to rescale the value into the interval $[0,1]$ and avoid the trouble of class imbalance in the validation set. Then, calculate the cost matrix $\bm{C}$:
\begin{equation} 
\bm{C} = {\bm S}'- \bm{I}
\label{cost_matrix}
\end{equation}
\noindent where $\bm{I}$ denotes the identity matrix. The diagonal of $\bm{C}$ can be negative, which is different from the general cost matrix: 
\begin{gather} 
c_{p,p} = s'_{p,p} -1 < 0 
,\\
c_{p,q} = s'_{p,q} > 0 \quad (p \neq q)
.\end{gather}
Next, update the generator matrix of labels:
\begin{equation} 
\bm{G} = (1 - \mu){\bm G}^*\! +\! \mu (\bm{I} - \epsilon \bm{C})
\label{update_generator_matrix}
\end{equation}
\noindent where the updating rate $\mu$ and the enhancement rate $\epsilon$ are two hyper-parameters, ${\bm G}^*$ is the old generator matrix before updating. When updating $\bm{G}$, the residual value (i.e. $(1-\mu)\bm{G}^*$) is to make the training process more stable. It can be proved easily that the properties described by Expression \eqref{generator_matrix_property_1}, \eqref{generator_matrix_property_2}, and \eqref{generator_matrix_property_3} are attached to the generator matrix $\bm{G}$ in Equation \eqref{update_generator_matrix}. Lastly, encode the labels of instances in the training set:
\begin{equation} 
{\bm L}^{\rm e}_{tr} = {\bm L}_{tr} \bm{G}
\label{encode_training_labels}
\end{equation}

\noindent where ${\bm L}_{tr}$ contains the one-hot labels of training instances. The whole BP algorithm can be briefly described as Algorithm \ref{the_algorithm}. 

Note that some details are omitted in Algorithm \ref{the_algorithm}, such as adjusting the learning rate dynamically, $\text{L}_1$ or $\text{L}_2$ regularization, etc. You can add them by yourself if you need to.

The mathematical derivation of enhancement encoding, which proves its effectiveness, is shown in the Appendices.

%%%%%%%%%%%%%%%%%%%%%%%%%%%%%%%%%%%%%%%%%%%%%%%%%%%%%%%%%%%%

% \clearpage
\section{Experiments}

\label{experiments}
In this section, we do some experiments to evaluate the performance of enhancement encoding and report the results in the end.

% \clearpage
\subsection{Experimental Setup}

\begin{wrapfigure}{0}{0.6\linewidth}
\centering
\includegraphics[width=\linewidth,trim = 60. 150. 60. 100, clip]{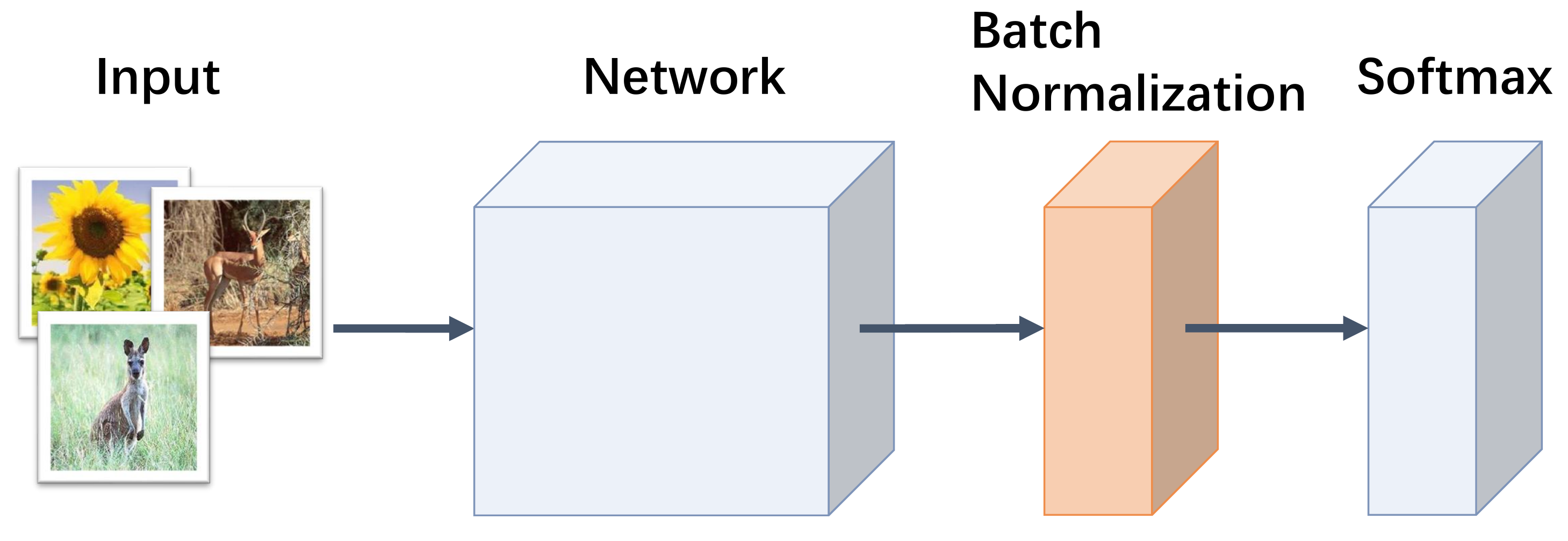}
\caption{We add a Batch Normalization layer before the Softmax. And then, the enhancement rate can be set as a higher value.}
\label{batchnormalization}
\end{wrapfigure}

In every experiment, we add a \emph{Batch Normalization layer} \cite{BatchNormalization} before the output activation function (i.e. \emph{Softmax}), as shown in Figure \ref{batchnormalization}. The Batch Normalization layer is very helpful to improve the performance of the enhancement encoding. When it is applied, the enhancement rate can be set as a higher value. 

\textbf{MNIST:} MNIST \cite{MNIST&LeNet-5} is a fairly basic standard data set, which is usually chosen to be the start for learning computer vision. It contains 7{,}000 handwriting digit images, 6{,}000 of which are for training and the rest are for tests. The original data set is balanced. To evaluate our algorithm, we manually unbalance the data in two ways: ``10\% of even classes'' and ``long-tailed distribution'' (see the Appendices). For the first way (10\% of even classes), we refer to the data distribution created by Khan et al. \cite{cosenCNN}. We choose LeNet-5 \cite{MNIST&LeNet-5} as the backbone net and make some alterations. We utilize \emph{ReLU} \cite{relu} as the hidden layers' activation function, utilize \emph{Softmax} as the output activation function, and replace the original \emph{Gaussian connection} layer with a fully connected layer.

\textbf{CIFAR-10-LT:} The standard CIFAR-10 \cite{cifar} is a balanced data set with 10 classes in total. There are 5{,}000 images for training and 1{,}000 images for tests per class. We choose its long-tailed version (CIFAR-10-LT) \cite{classBalancedLoss} with the imbalance factor of 100 to evaluate our approach. The imbalance factor \cite{classBalancedLoss} describes the imbalance degree of the data set. It can be computed by $\left.\mathop{\max}\limits_p n_p\right./\left.\mathop{\min}\limits_p n_p\right.$,  where $n_p$ denotes the number of instances belonging to class $p$. We randomly select 5 instances per class as the validation set. We choose the ResNet-32 \cite{resnet} as the backbone net. 

\textbf{Caltech 101:} Caltech 101 \cite{caltech101} is an originally imbalanced data set. It contains 8{,}677 images belonging to 101 classes (exclude the class named ``BACKGROUND\_Google''). There are about 30$\thicksim$800 images per class and the imbalance factor of Caltech 101 is 25.81 ($\text{IF}=800/31 =25.81$). We randomly split the data set into three parts: a validation set containing 5 images per class, a test set containing 20 images per class, and a training set containing the rest of the data. The resolution of these images is roughly 300$\times$200 and we resize them to 224$\times$224. The backbone net is ResNet-34.

We compare the performances of enhancement encoding and one-hot encoding in three different situations (cross-entropy loss, mean square error, and focal loss \cite{focalLoss}). Focal loss of instance $i$ can be expressed as:
\begin{equation}
    J_i =-\sum_{q=0}^{N-1} \alpha_q\, l_{i,q} (1-y_{i,q})^\gamma \log y_{i,q}
    \label{focalLoss}
\end{equation}
\noindent where $\alpha_q$ denotes the class weight and $\gamma$ denotes the focusing parameter. Because the enhancement labels re-weight the instances of different classes, we set all class weights ($\alpha_q$) as ``$1.0$''.

As mentioned in \eqref{generator_matrix_property_1} and \eqref{generator_matrix_property_2}, the enhancement encoding technique attaches both re-weighting and cost-sensitiveness to the training process. So we also evaluate the incomplete enhancement labels by setting the non-diagonal elements of generation matrix $\bm{G}$ (see Section \ref{implementation}) as ``0'' (marked with ``\emph{re-weighting}'') or setting the diagonal elements as ``1'' (marked with ``\emph{cost-sensitiveness}'').

\begin{table}[!htb]
\caption{Test Top-1 Accuracy on Imbalanced MNIST Data Set (\%)}
\label{results_of_MNIST}
\centering
\renewcommand{\arraystretch}{1.5}
\small
\begin{tabular}{l|c|c c c c}
\hline
\makecell[l]{{\bf Imbalance}\\{\bf Settings}} & {\bf Losses} & \makecell{Baseline (One-\\Hot Encoding)} & \makecell{Re-\\Weighting} & \makecell{Cost-\\Sensitiveness} & \makecell{Enhancement\\Encoding} \\
    % &   & Hot Encoding) & Weighting & Sensitiveness & Encoding \\
\hline
\multirow{3}*{\makecell[l]{10\% of Even\\Classes}} & MSE & 97.82$\pm$0.03 & 97.92 & 97.89 & {\bf 97.99}$\pm$0.06  \\
                                    & CE  & 97.75$\pm$0.04 & 97.85 & 97.91 & {\bf 98.00}$\pm$0.03 \\
                                    & FL ($\gamma=2.0$, $\alpha_q=1.0$) & 98.06$\pm$0.04 & 98.21 & 98.32 & {\bf 98.54}$\pm$0.13  \\
\hline
\multirow{3}*{\makecell[l]{Long-Taile\\Distribution}} & MSE & 92.12$\pm$0.56 & 92.56 & 92.61 & {\bf 92.78}$\pm$0.26 \\ 
                                    & CE & 89.14$\pm$0.13 & 90.95 & 90.65 & {\bf 91.43}$\pm$0.27 \\
                                    & FL ($\gamma=2.0$, $\alpha_q=1.0$) & 91.56$\pm$0.32 & 91.80 & 91.87 & {\bf 93.34}$\pm$0.61 \\
\hline
\end{tabular}
\end{table}

\subsection{Results}

\begin{wraptable}{0}{0.72\linewidth}
\caption{Test Performances on CIFAR-10-LT Data Set (\%)}
\label{results_of_cifar_10_lt}
\centering
\renewcommand{\arraystretch}{1.3}
\small
\begin{tabular}{l|c|c c}
\hline
{\bf Losses} & {\bf Methods} & \makecell[c]{{\bf Top-1}\\{\bf Accuracy}} & \makecell[c]{{\bf Minority}\\{\bf Accuracy}}\\
\hline
\multirow{4}*{MSE} & Baseline (One-Hot Encoding) & 67.15$\pm$0.48 & 49.84$\pm$1.01 \\ \cdashline{2-4}
                  & Re-Weighting & 67.20 & 50.62 \\ \cdashline{2-4}
                  & Cost-Sensitiveness & 67.39 & 50.22 \\ \cdashline{2-4}
                  & Enhancement Encoding & {\bf 67.89}$\pm$0.58 & {\bf 51.59}$\pm$1.17 \\
\hline
\multirow{4}*{CE} & Baseline (One-Hot Encoding) & 65.23$\pm$0.75 & 47.34$\pm$1.56 \\ \cdashline{2-4}
                  & Re-Weighting & 69.55 & 52.38 \\ \cdashline{2-4}
                  & Cost-Sensitiveness & 65.98 & 49.48 \\ \cdashline{2-4}
                  & Enhancement Encoding & {\bf 69.89}$\pm$0.89 & {\bf 55.75}$\pm$1.73 \\
\hline
\multirow{4}*{\makecell[l]{FL\\($\gamma=1.0,$\\$\alpha_q=1.0$)}} & Baseline (One-Hot Encoding) & 67.33$\pm$0.35 & 50.34$\pm$0.78 \\ \cdashline{2-4}
                  & Re-Weighting & 68.10 & 51.54 \\ \cdashline{2-4}
                  & Cost-Sensitiveness & 69.49 & 55.18 \\ \cdashline{2-4}
                  & Enhancement Encoding & {\bf 69.94}$\pm$0.79 & {\bf 56.38}$\pm$1.24 \\
\hline
\end{tabular}
\end{wraptable}

We define the test \emph{minority accuracy} as the top-1 accuracy evaluated on test instances belonging to minority classes. We regard the minority classes as the $N_{min}$ classes with the fewest instances in the training set.

\begin{figure}[!htb]
\centering
\hspace{-1.5mm}
\subfloat[Baseline]
    {\label{baseline_scm}
     \centering
     \includegraphics[width=0.243\textwidth,trim = 85 5 101 15, clip]{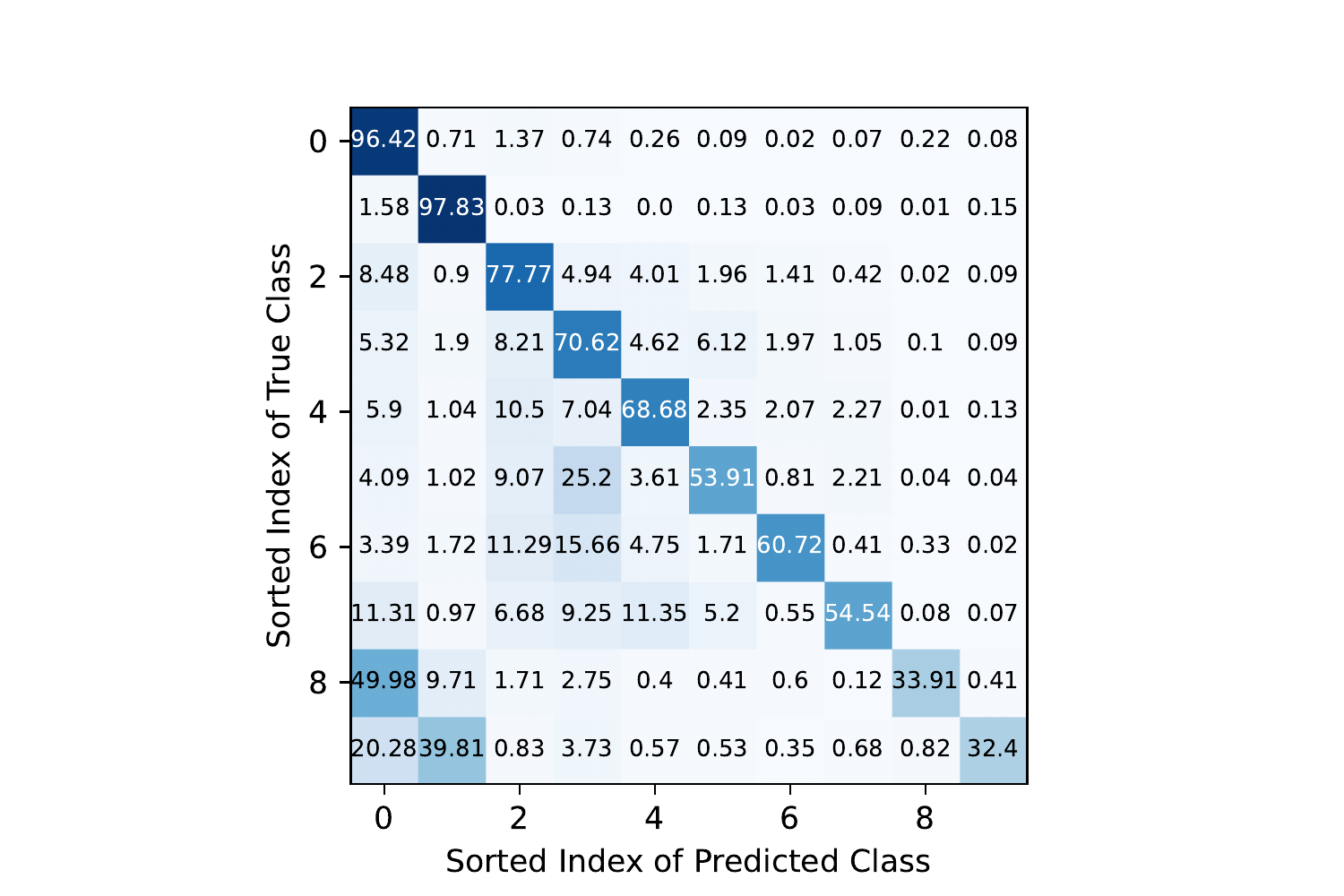}
    }
\hspace{-1.5mm}
\subfloat[Enhancement\\Encoding]
    {\label{enhancement_encoding_scm}
     \centering
     \includegraphics[width=0.243\textwidth,trim = 85 5 101 15, clip]{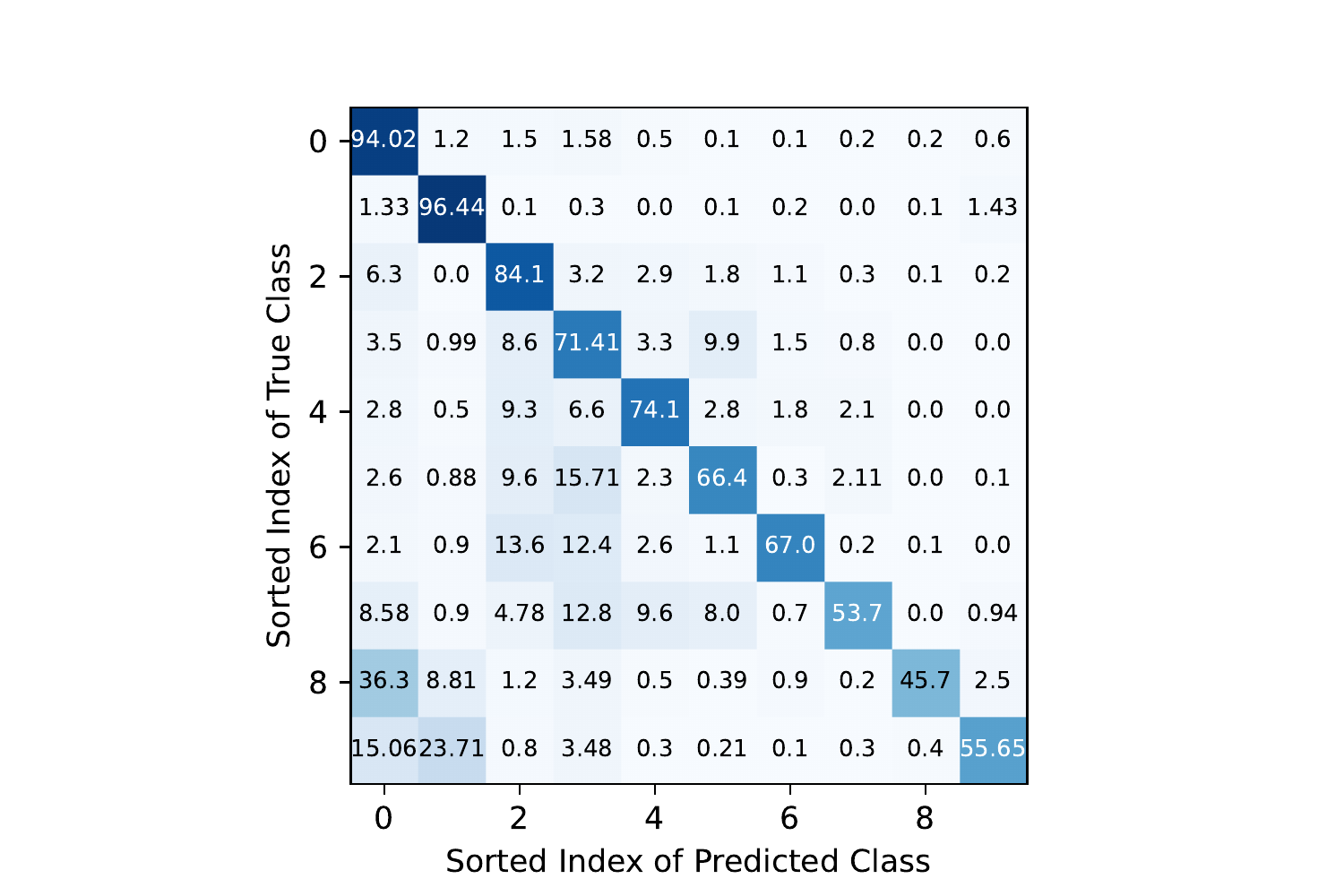}
    }
\hspace{-1.5mm}
\subfloat[Re-Weighting]
    {\label{re_weighting_scm}
     \centering
     \includegraphics[width=0.243\textwidth,trim = 85 5 101 15, clip]{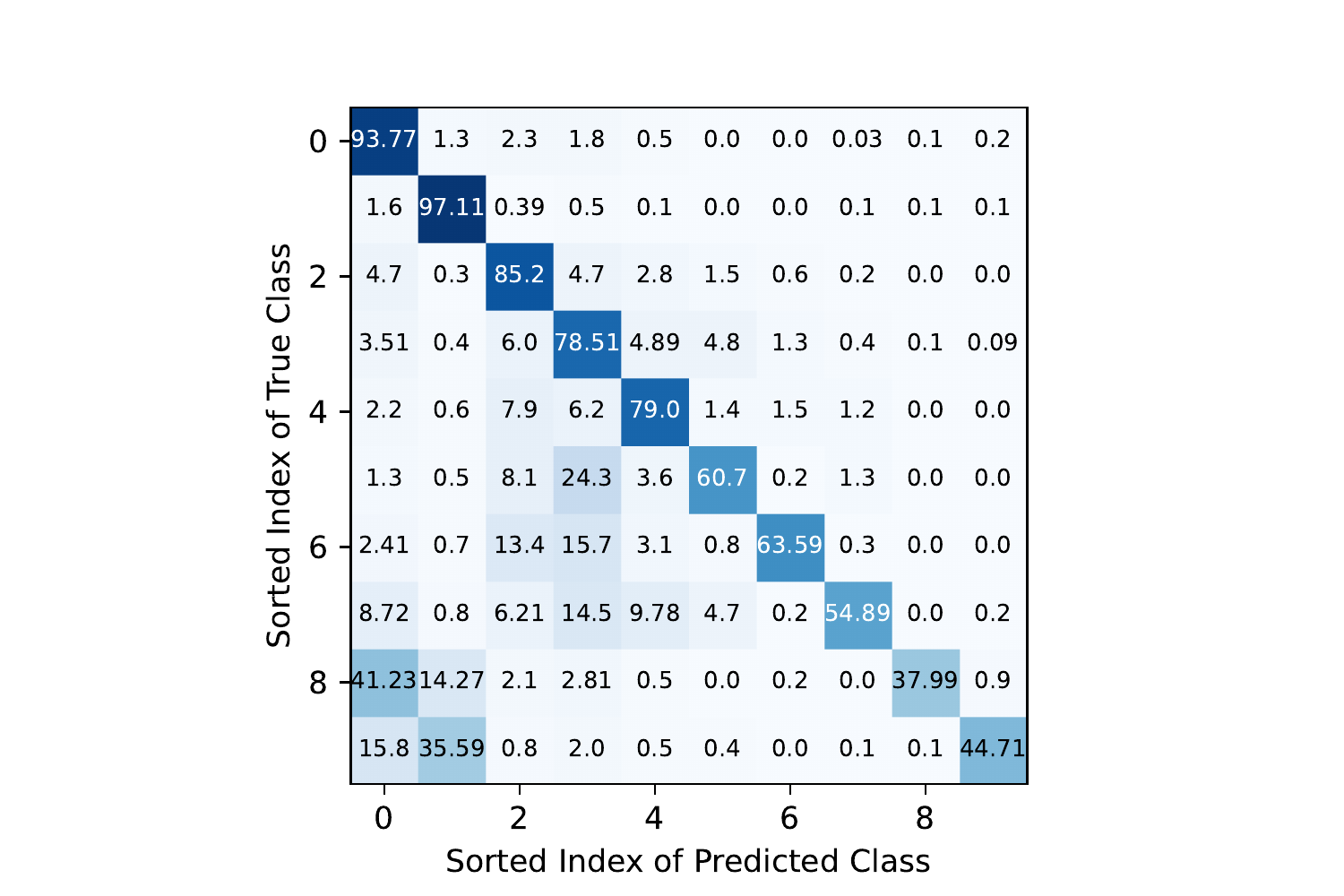}
    }
\hspace{-1.5mm}
\subfloat[Cost-Sensitiveness]
    {\label{cost_sensitiveness_scm}
     \centering
     \includegraphics[width=0.243\textwidth,trim = 85 5 101 15, clip]{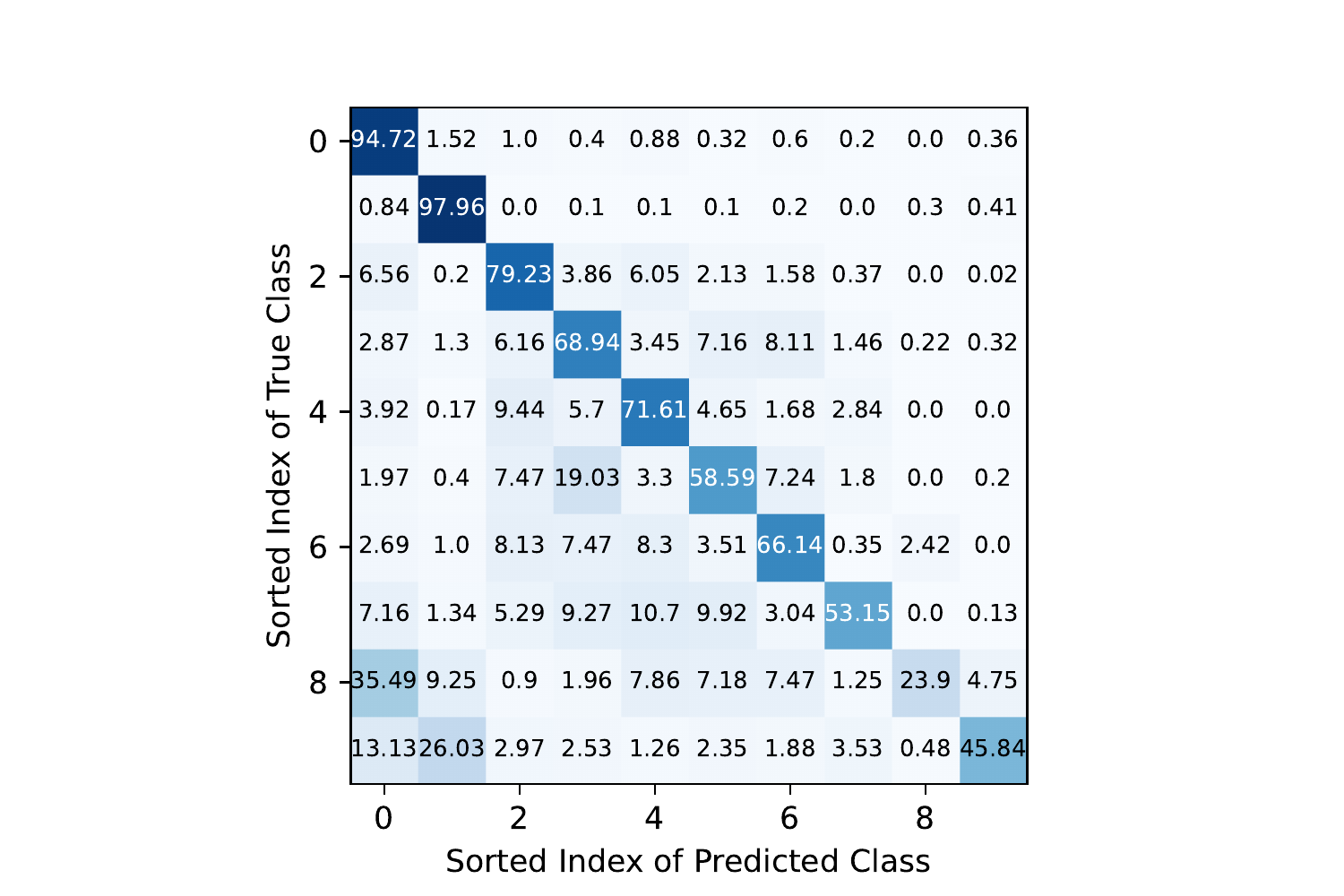}
    }
\hspace{-1.5mm}
\caption{The row-normalized test soft-confusion matrices of networks which are trained with cross-entropy on CIFAR-10-LT. The percent signs are omitted.}
\label{cifar_10_lt_scm}
\end{figure}

\begin{figure}[!htb]
\centering
% \hspace{-1.5mm}
\subfloat[Baseline]
    {\label{baseline_t-sne}
     \centering
     \includegraphics[width=0.23\textwidth,trim = 48 48 44 47, clip]{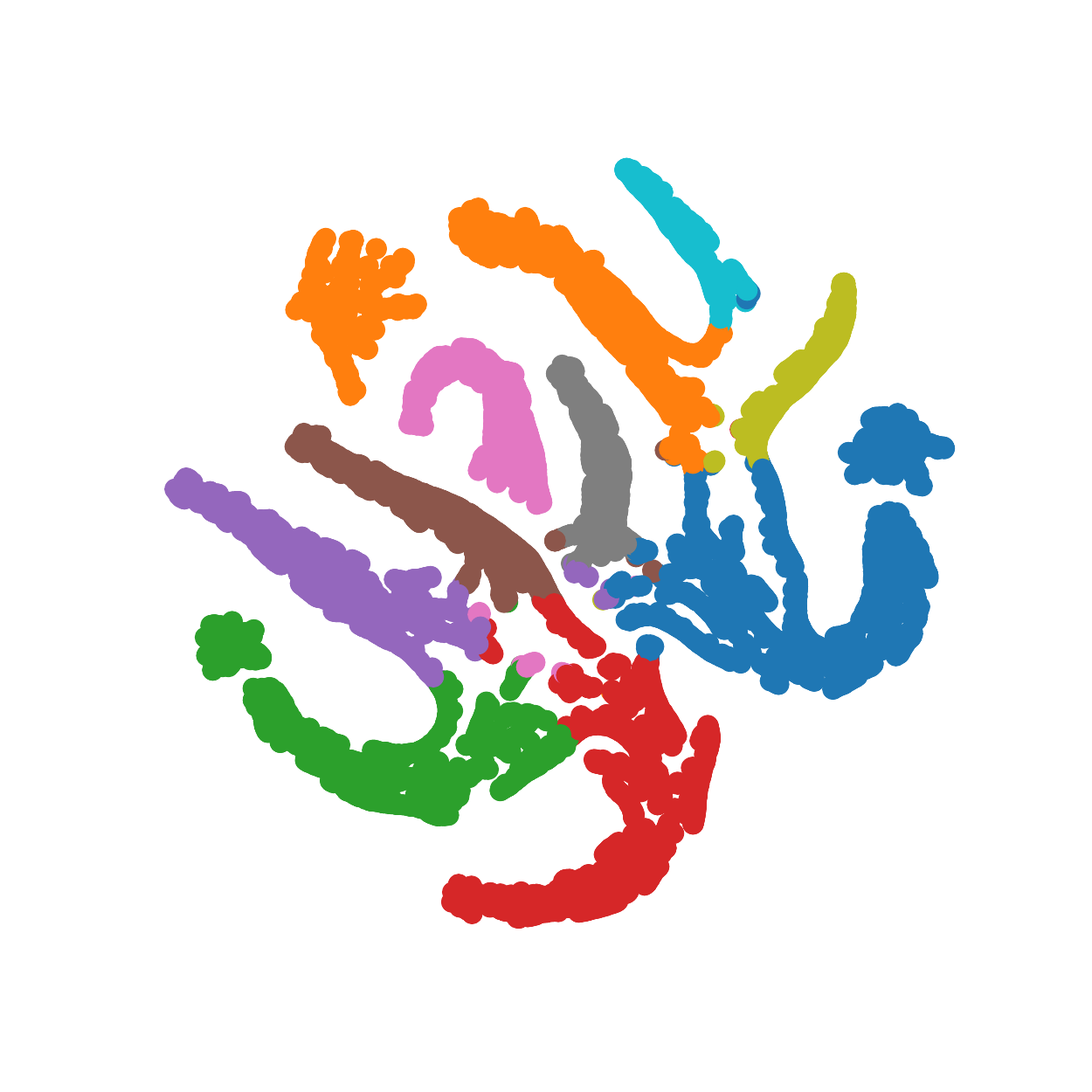}
    }
\hspace{1cm}
\subfloat[Enhancement Encoding]
    {\label{enhancement_encoding_t-sne}
     \centering
     \includegraphics[width=0.23\textwidth,trim = 48 48 44 47, clip]{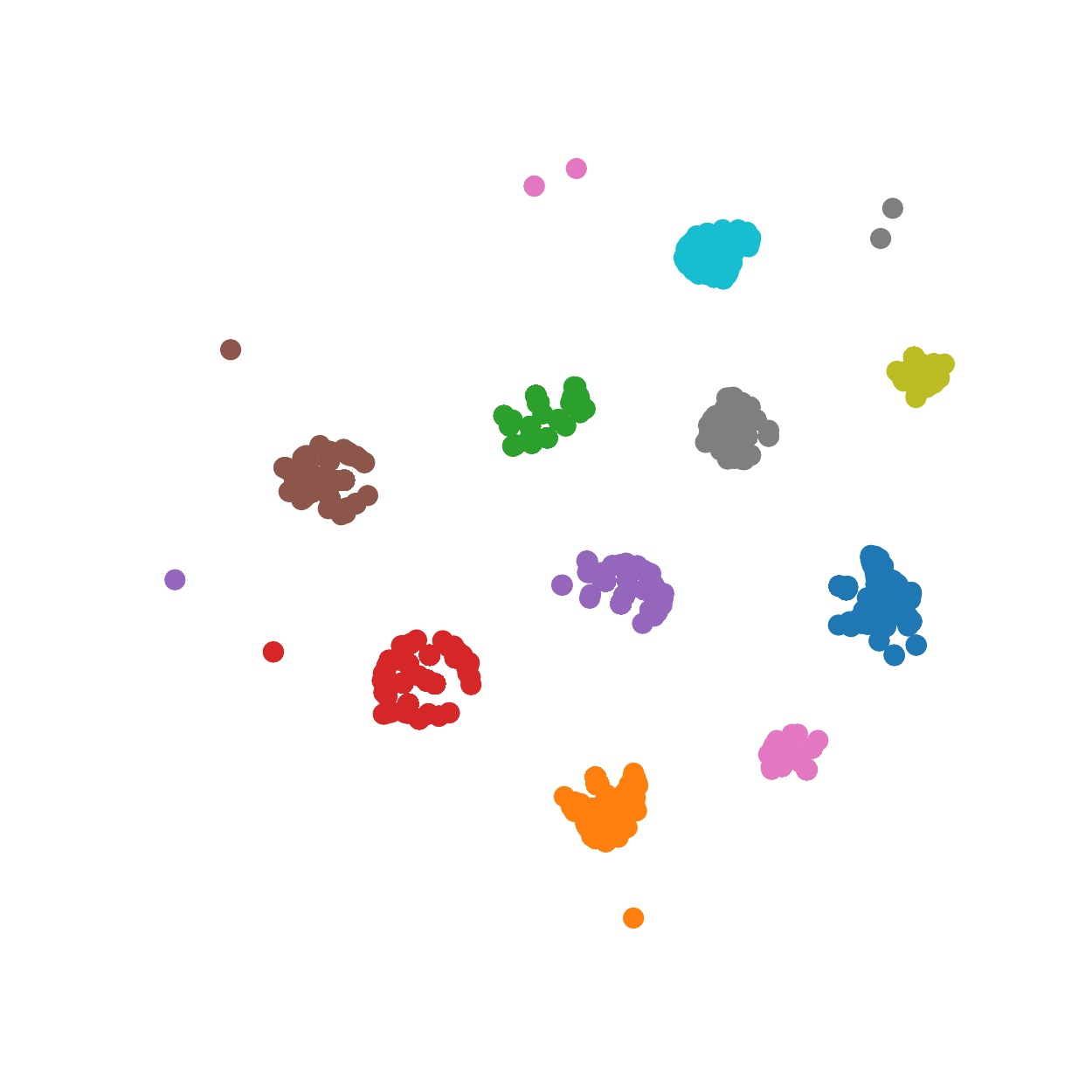}
    }
\hspace{1cm}
\subfloat
    {\centering
     \includegraphics[height=0.23\textwidth,trim = 21 48 20 47, clip]{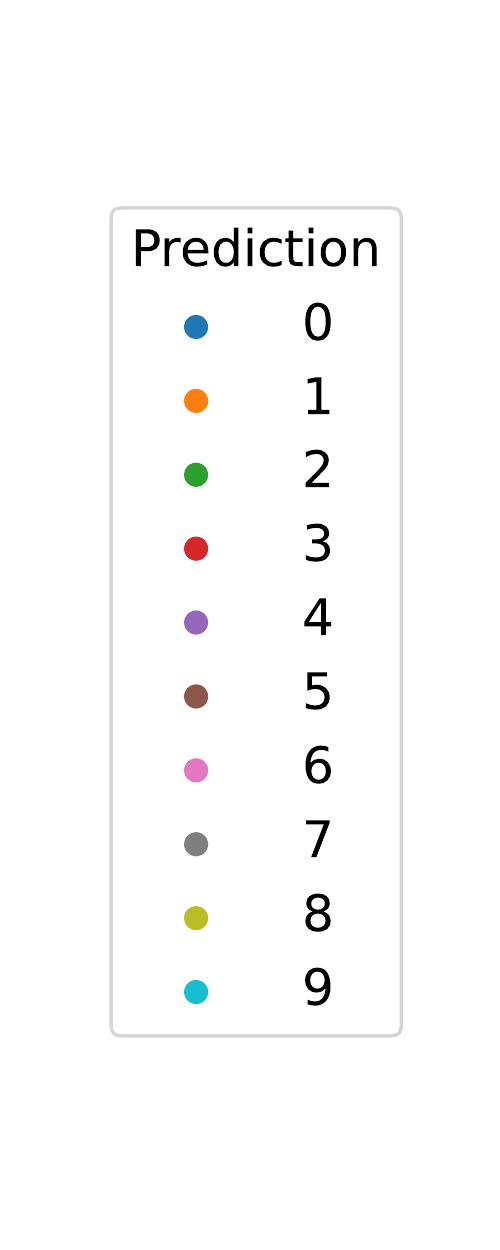}
    }
\caption{The t-SNE visualization of classification results on the test set of CIFAR-10-LT. In Fig. (b), Samples with the same prediction result are more concentrated}
\label{t-sne}
\end{figure}

For MNIST, we report the top-1 accuracy in Table \ref{results_of_MNIST}. For CIFA-10-LT,  we report the top-1 accuracy and the minority accuracy ($N_{min} = 5$) in Table \ref{results_of_cifar_10_lt}. And for Caltech 101, top-1 accuracy, top-5 accuracy, and minority accuracy ($N_{min} = 50$) are shown in Table \ref{results_of_caltech101}. Obviously, with any one of the three losses, the enhancement encoding always performs better than the one-hot encoding. Particularly, the minority accuracy of enhancement encoding is much higher in all experiments. As shown by the soft-confusion matrices in Figure \ref{cifar_10_lt_scm}, our method may decline the performance on some majority classes (class 0 and class 1) a little bit, but it improves the performance on minority classes (class 5-9) a lot. In Figure \ref{t-sne}, we show the t-SNE visualization \cite{t-sne} of the network's outputs. It can be found that enhancement encoding makes samples with the same prediction result more concentrated. We run the programs of baseline and enhancement encoding 5 times per situation and report the error bars (Mean$\pm$SD) in Table \ref{results_of_MNIST}, Table \ref{results_of_cifar_10_lt}, and Table \ref{results_of_caltech101}.

\textbf{Ablation Studies:} We evaluate the two incomplete enhancement encoding methods (re-weighting and cost-sensitive) and show the results in Table \ref{results_of_MNIST}, Table \ref{results_of_cifar_10_lt}, and Table \ref{results_of_caltech101}. It can be observed that the two types of encoding also perform better than the baseline in most situations. Note the four elements located at the bottom left corner of the normalized soft-confusion matrices ($s_{8,0}'$, $s_{8,1}'$, $s_{9,0}'$, and $s_{9,1}'$) in Figure \ref{cifar_10_lt_scm}, the cost-sensitiveness is lower, overall. It illustrates that the cost-sensitiveness part of our method is very helpful to mitigate the adverse impact that the majority data overwhelms the minority. 

\begin{table}[!htb]
\caption{Test Performances on Caltech 101 Data Set (\%)}
\label{results_of_caltech101}
\centering
\renewcommand{\arraystretch}{1.3}
\setlength\tabcolsep{5.7pt}
\small
\begin{tabular}{l|c|c c c}
\hline
{\bf Losses} & {\bf Methods} & \makecell[c]{\bf Top-1 Accuracy} & \makecell[c]{\bf Top-5 Accuracy} &\makecell[c]{\bf Minority Accuracy}\\
\hline
\multirow{4}*{MSE} & Baseline (One-Hot Encoding) & 54.66$\pm$0.48  & 73.55$\pm$0.37 & 37.32$\pm$0.33\\ \cdashline{2-5}
                  & Re-Weighting & 57.03 & 75.05 & 41.10 \\ \cdashline{2-5}
                  & Cost-Sensitiveness & 54.75 & 73.22 & 38.60 \\ \cdashline{2-5}
                  & Enhancement Encoding & {\bf 57.16}$\pm$0.87 & {\bf 75.82}$\pm$0.86 & {\bf 41.28}$\pm$1.22 \\
\hline
\multirow{4}*{CE} & Baseline (One-Hot Encoding) & 58.30$\pm$0.93 & 77.45$\pm$1.56 & 43.14$\pm$0.81 \\ \cdashline{2-5}
                  & Re-Weighting & 60.50 & {\bf 80.15} & 46.90 \\ \cdashline{2-5}
                  & Cost-Sensitiveness & 58.81 & 78.42 & 42.70\\ \cdashline{2-5}
                  & Enhancement Encoding & {\bf 61.48}$\pm$1.07 & 79.50$\pm$0.41 & {\bf 47.54}$\pm$1.69\\
\hline
\multirow{4}*{\makecell[l]{FL\\($\gamma=0.5,$\\$\alpha_q=1.0$)}} & Baseline (One-Hot Encoding) & 57.28$\pm$1.38 & 77.46$\pm$0.52 & 42.34$\pm$1.61\\ \cdashline{2-5}
                  & Re-Weighting & 58.17 & 76.44 & 42.70 \\ \cdashline{2-5}
                  & Cost-Sensitiveness & 59.85 & {\bf 78.86} & 46.10\\ \cdashline{2-5}
                  & Enhancement Encoding & {\bf 60.27}$\pm$1.11 & 78.80$\pm$0.88 & {\bf 46.82}$\pm$1.43\\
\hline
\end{tabular}
\end{table}

%%%%%%%%%%%%%%%%%%%%%%%%%%%%%%%%%%%%%%%%%%%%%%%%%%%%%%%%%%%%

\section{Related Work}

Researches on imbalanced learning can be divided into two categories: data-level methods and algorithm-level methods \cite{studyOfCI}. The former is to change the distribution of the training data and the latter is to change the strategy of training.

\subsection{Data-Level Methods}

Data-level methods include oversampling and undersampling.

\textbf{Oversampling:} This kind of method refers to increasing the number of minority instances to balance the data set. The simplest approach is to repeat the minority data during training, but in this way, the decision region of the minority class may be shrunk, which goes against the expectation \cite{smote}. Another idea for oversampling is to artificially produce new instances. Ha and Bunke \cite{perturbationMethod} adjusted the original data slightly to get new training data. A more representative method is SMOTE, which is proposed by Chawla et al. \cite{smote}. SMOTE is to linearly synthesize instances between a sample and its $k$ neighbors with a little random perturbation. Based on the research of \cite{smote}, some verities of SMOTE have been proposed, such as SMOTEBoost (Chawla et al. \cite{smoteBoost}), borderline-SMOTE (Han et al. \cite{bSmote}), k-means SMOTE (Last \cite{kSmote}), etc.

\textbf{Undersampling:} In contrast to oversampling, undersampling means deleting some majority instances to keep the training set balanced. It is generally believed that undersampling is not a good option for imbalanced learning because it strips out some information from the training set, but Drummond and Holte \cite{usBeatsOs} found that undersampling may perform better than oversampling in some cases. There are some undersampling methods: Tomek \cite{tomekLinks} removed all the \emph{Tomek Links} during training, Zhang and Mani \cite{NearMiss} designed the NearMiss algorithm, and Liu et al. \cite{EasyEnsemble} proposed EasyEnsemble.

\subsection{Algorithm-Level Methods}

There are three kinds of algorithm-level methods for imbalanced classification. Note that these methods are not in conflict and sometimes they can work together.

\textbf{Thresholding:} In most classification tasks, the decision result is the class whose prediction probability is the highest (multi-classification) or higher than the threshold (binary classification). Thresholding is to adjust the decision threshold or rescale the prediction probability. Hern\'{a}ndez-Orallo et al. \cite{translatingThresholdChoice} systematically compared several threshold choice methods with different metrics and described how to pick one. Sheng and Ling \cite{thresholdingCostSensitive} implemented cost-sensitiveness via thresholding. Collell et al. \cite{thresholdBagging} combined thresholding and the bagging algorithm (\emph{PT-bagging}).

\textbf{Cost-Sensitive Learning:} Unlike general machine learning algorithms aiming to minimize classification error, cost-sensitive learning minimizes the cost of misclassification. With the larger cost of classifying a minority instance into a majority class, the classifier can reduce the misclassification of the minority data. Masnadi-Shirazi and Vasconcelos \cite{cosenSVM} combined the SVM and cost-sensitive learning. Khan et al. \cite{cosenCNN} altered the loss functions to make neural network cost-sensitive. Menon et al. \cite{logitAdjustment} proposed a very similar approach. The main difference is that the cost in the former is learned from the data and the latter is based on the prior probability.

\textbf{Re-weighting:} This kind of method can be considered as setting different weights of loss function for the majority or minority data. Lin et al. \cite{focalLoss} designed \emph{Focal Loss} (see Equation \eqref{focalLoss}), which calculates the classification weights with outputs of the network. Cui et al. \cite{classBalancedLoss} re-weighted instances belonging to different classes based on the \emph{effective number}. Cao et al. \cite{LDAM} proposed the \emph{Label-Distribution-Aware Margin Loss}. Ren et al. \cite{learningToReweight} and Jamal et al. \cite{RethinkingCB} separately designed two meta-learning algorithms to get the class weights.

%%%%%%%%%%%%%%%%%%%%%%%%%%%%%%%%%%%%%%%%%%%%%%%%%%%%%%%%%%%%

\section{Conclusion and Future Work}
\label{conclusion}
In this paper, we introduce a novel enhancement encoding, which combines re-weighting and cost-sensitiveness. It provides an entirely new idea to address the class imbalance problem by changing the way to encode training labels. We also present the soft-confusion matrix, which reduces the validation data and extra computation for enhancement encoding. We evaluate the performances of enhancement encoding and separately study the effects of re-weighting and cost-sensitive. The results show that the enhancement encoding of labels effectively improves the performances of the neural network on the imbalanced classification, especially for the minority data.

At present, there are still some limitations of our approach and we hope to settle them in the future work. Firstly, We evaluate the enhancement encoding with some classic networks (LeNet-5 \cite{MNIST&LeNet-5} and ResNet \cite{resnet}). Experiments on emerging models (ViT \cite{vit}, External Attention Transformer \cite{eat}, etc.) are lacking. How much the enhancement encoding improves the performance of these new methods is still not clear. Secondly, we at present find the enhancement rate $\epsilon$ and the updating rate $\mu$ (see Equation \eqref{update_generator_matrix}) by comparing the validation accuracy, which takes too much time and energy. Next, we are going to find a more efficient way to fine-tune the two hyper-parameters and explore if the hyper-parameters can be dynamic during training.

%%%%%%%%%%%%%%%%%%%%%%%%%%%%%%%%%%%%%%%%%%%%%%%%%%%%%%%%%%%%

%Bibliography
\clearpage

\clearpage
\appendix
\part*{Appendices}

\section{Mathematical Derivation on BP with Enhancement Encoding}

\label{MathematicalDeduction}

In this appendix, we are going to talk about how the enhancement encoding influences the BP algorithm and show how it works effectively for imbalanced learning with cross-entropy loss and mean square error loss.

\subsection{Cost-sensitiveness and Re-weighting}

According to Equation~\eqref{oh_label} and Equation \eqref{encode_one_training_label}, it can be inferred that if the instance $i$ belongs to class $p$, the enhancement label of instance $i$ is equal to the $p$-th row of generator matrix $\bm{G}$:
\begin{equation}
    {\bm l}^{\rm e}_i = {\bm g}_p \quad (i \in {\bf D}_p)
.\end{equation}
Then, some generator matrix elements' properties described by Expression \eqref{generator_matrix_property_1}, \eqref{generator_matrix_property_2}, and \eqref{generator_matrix_property_3} are also attached to the enhancement labels.

That is if the instance $i$ belongs to class $p$ ($i \in {\bf D}_p$), then:
\begin{gather}
    l^{\rm e}_{i,p} >1 
,\\
    l^{\rm e}_{i,q} <0 \quad (q \neq p)
,\\
    \sum_{q=0}^{N-1} l^{\rm e}_{i,q} =1
    .\label{sum_of_e_label}
\end{gather}
To simplify the analysis, let $\mu$ in Equation \eqref{update_generator_matrix} be ``$1$'':
\begin{equation} 
\bm{G} = \bm{I}- \epsilon \bm{C}
% \label{update_generator_matrix}
.\end{equation}
Note the elements of one-hot labels in Equation \eqref{oh_label}, then:
\begin{gather}
    \begin{aligned}
        l^{\rm e}_{i,p} &= g_{p,p}\\
                  &= 1-\epsilon c_{p,p}\\
                  &= l_{i,p} - \epsilon c_{p,p} \quad (i \in {\bf D}_p)
    ,\end{aligned}
\\\notag\\
    \begin{aligned}
        l^{\rm e}_{i,q} &= g_{p,q}\\
                  &= -\epsilon c_{p,q}\\
                  &= l_{i,q} - \epsilon c_{p,q} \quad (i \in {\bf D}_p\,,\ p\neq q)
    .\end{aligned}
\end{gather}
So: 
\begin{equation}
    \begin{aligned}
        l^{\rm e}_{i,q} - l_{i,q} &= -\epsilon c_{p,q}\\
            % & \\
            &= 
            \left\{
                \begin{aligned}
                    &-\epsilon c_{p,p} >0, &&p=q (\text{re-weighting}) \\
                    &-\epsilon c_{p,q} <0, &&p\neq q (\text{cost-sensitiveness})
                .\end{aligned}
            \right. \\
    \end{aligned}
    \label{defference_of_labels}
\end{equation}
Equation \eqref{defference_of_labels} shows how the enhancement labels implement cost-sensitiveness and re-weighting.

\subsection{Situation with Cross-Entropy Loss}

\label{situation_with_CE}
When dealing with classification, cross-entropy is usually utilized as the network's loss function. It can be described as follows: 
\begin{equation}
    J_i =-\sum_{q=0}^{N-1} l_{i,q} \log y_{i,q}\,
    .\label{CE_loss}
\end{equation}
Let the base number of logarithms in Equation \eqref{CE_loss} be the natural constant ``$e$'', then the partial derivative of cross-entropy with respect to $y_{i,q}$ is:
\begin{equation}
    \frac{\partial J_i}{\partial y_{i,q}}=-\frac{l_{i,q}}{y_{i,q}}
    .\label{partial_derivative_of_CE_y}
\end{equation}
The Softmax activation of the output layer is another critical function for analysis, whose expression is:
\begin{equation}
    y_{i,q}=\frac{e^{z_{i,q}}}{\sum_{j=0}^{N-1} e^{z_{i,j}}}
\end{equation}
\noindent where $z_{i,k}$ denotes the weighted sum of the output layer's inputs. Softmax activation can rescale the network's outputs into the interval $(0,1)$ and let their sum be ``$1$''. This attribute of Softmax can make the network predict probability very well. The partial derivative is:
\begin{equation}
    \begin{aligned}
        \frac{\partial y_{i,q}}{\partial z_{i,k}} 
            % &= \frac{
            % \frac{\partial e^{z_{i,q}}}{\partial z_{i,k}}\left[\sum_{j=1}^N e^{z_{i,j}}\right]-e^{z_{i,q}} e^{{z_{i,k}}}{\left[\sum_{j=1}^N e^{z_{i,j}}\right]^{2}}\\
            &=\frac{\frac{\partial e^{z_{i,q}}}{\partial z_{i,k}}\left[\sum_{j=0}^{N-1} e^{z_{i,j}}\right]-e^{z_{i,q}} e^{z_{i,k}}}{\left[\sum_{j=0}^{N-1} e^{z_{i,j}}\right]^{2}}\\
            &=\frac{\frac{\partial e^{z_{i,q}}}{\partial z_{i,k}}}{\sum_{j=0}^{N-1} e^{z_{i,j}}}-y_{i,q} \  y_{i,k}\\
            &=
                \left\{
                    \begin{aligned}
                        &\ y_{i,k}-y_{i,k}^{2}\,, && q=k \\ 
                        &-y_{i,q} \  y_{i,k}\,, && q \neq k
                    .\end{aligned}
                \right.
    \end{aligned}
\end{equation}

Go further:
\begin{equation}
    \begin{aligned}
        \frac{\partial J_i}{\partial z_{i,k}}
        &=\sum_{q=0}^{N-1} \frac{\partial J_i}{\partial y_{i,q}} \frac{\partial y_{i,q}}{\partial z_{i,k}}\\
        &=\frac{\partial J_i}{\partial y_{i,k}}(y_{i,k}-y_{i,k}^{2})-\sum_{q \neq k} \frac{\partial J_i}{\partial y_{i,q}} y_{i,q} \  y_{i,k}\\
        &=\frac{\partial J_i}{\partial y_{i,k}} y_{i,k} -\sum_{q=0}^{N-1} \frac{\partial J_i}{\partial y_{i,q}} y_{i,q} \  y_{i,k}\\
        &=-\frac{l_{i,k}}{y_{i,k}} y_{i,k} +\sum_{q=0}^{N-1} \frac{l_{i,q}}{y_{i,q}} y_{i,q} \  y_{i,k}\\
        &=- l_{i,k} + \sum_{q=0}^{N-1} l_{i,q} \  y_{i,k}\\
        &=y_{i,k} - l_{i,k}
        \qquad \qquad \rhd \text{Eq.~\eqref{oh_label}}
    .\end{aligned} 
    \label{partial_derivative_of_CE_z}
\end{equation}
Obviously, enhancement labels also satisfy Equation \eqref{CE_loss} and \eqref{partial_derivative_of_CE_y}, so $l_{i,k}$ in Equation \eqref{partial_derivative_of_CE_z} can be replaced with $l^{\rm e}_{i,k}$:
\begin{equation}
    \begin{aligned}
        \frac{\partial J^{\rm e}_i}{\partial z_{i,k}}
        &=-l^{\rm e}_{i,k} + \sum_{q=0}^{N-1} l^{\rm e}_{i,q} \  y_{i,k}\\
        &=y_{i,k} - l^{\rm e}_{i,k}
        \qquad \qquad \rhd \text{Eq.~\eqref{sum_of_e_label}}
    .\end{aligned}
\end{equation}
When updating the network's parameters with gradient descent, the direction of updating vector $\Delta {\bm \theta}$ on the loss surface is opposite to the gradient of $J$:
\begin{equation}
    \Delta {\bm \theta} = -\eta \nabla_{\bm \theta} J 
.\end{equation}
Suppose instance $i$ belongs to class $p$ ($i \in {\bf D}_p$), the difference between BPs with and without enhancement labels can be described by:
\begin{equation}
    \begin{aligned}
        \text{DBP}_{i,k} 
            &= \left(-\frac{\partial J^{\rm e}_i}{\partial z_{i,k}} \right) - \left(-\frac{\partial J_i}{\partial z_{i,k}} \right)\\
            &= (l^{\rm e}_{i,k}-y_{i,k}) - (l_{i,k}-y_{i,k})\\
            &= l^{\rm e}_{i,k} - l_{i,k}\\
            &= -\epsilon c_{p,k} \qquad \qquad \rhd \text{Eq.~\eqref{defference_of_labels}}\\
            % & \\
            &= 
            \left\{
                \begin{aligned}
                    &-\epsilon c_{p,p} = \epsilon(1-s'_{p,p}) > 0, &&p=k\\
                    &-\epsilon c_{p,k} = -\epsilon s'_{p,k} < 0, &&p\neq k
                .\end{aligned}
            \right. 
    \end{aligned} 
    \label{DBP_CE}
\end{equation}
Equation \eqref{DBP_CE} means that compared with one-hot labels, the BP with enhancement labels makes $z_{i,p}$ increase more and $z_{i,k}$ ($k\neq p$) decrease more.

Suppose there is an instance $a$ belonging to a minority class $u$ ($a \in {\bf D}_u$) and an instance $b$ belonging to a majority class $v$ ($b \in {\bf D}_v$). In general:
\begin{gather}
    s'_{u,u}<s'_{v,v}\ 
,\\
    s'_{u,v}>s'_{v,u}\ 
.\end{gather}
Then:
\begin{gather}
    \text{DBP}_{a,u}>\text{DBP}_{b,v}>0 \label{DBP_au__DBP_bv}
,\\
    \text{DBP}_{a,v}<\text{DBP}_{b,u}<0 \label{DBP_av__DBP_bu}
.\end{gather}
Expression \eqref{DBP_au__DBP_bv} means $z_{a,u}$ increases more than $z_{b,v}$ and Expression \eqref{DBP_av__DBP_bu} means $z_{a,v}$ decreases more than $z_{b,u}$. In other words, the minority classes are attached with heavier class weights and higher misclassification costs than the majority classes.

\subsection{Situation with Mean Square Error Loss}

Mean square error (MSE) loss is another useful loss function in machine learning. It can be described as follows:
\begin{equation}
    J_i =\frac{1}{N} \sum_{q=0}^{N-1} (y_{i,q}-l_{i,q})^2
    .\label{MSE_loss}
\end{equation}
The partial derivative of MSE with respect to $y_{i,q}$ is:
\begin{equation}
    \frac{\partial J_i}{\partial y_{i,q}}=\frac{2}{N}(y_{i,q}-l_{i,q})
    .\label{partial_derivative_of_MSE_y}
\end{equation}
And the partial derivative of MSE with enhancement labels is:
\begin{equation}
    \frac{\partial J^{\rm e}_i}{\partial y_{i,q}}=\frac{2}{N}(y_{i,q}-l^{\rm e}_{i,q})
    % \label{partial_derivative_of_MSE_y}
.\end{equation}
When analyzing the network with MSE loss, Softmax activation can complicate the derivation. So, we define $\text{DBP}_{i,k}$ ($i \in {\bf D}_p$) with MSE based on partial derivative with respect to $y_{i,q}$\,:
\begin{equation}
    \begin{aligned}
        \text{DBP}_{i,k} 
            &= \left(-\frac{\partial J^{\rm e}_i}{\partial y_{i,k}} \right) - \left(-\frac{\partial J_i}{\partial y_{i,k}} \right)\\
            &= \frac{2}{N}(l^{\rm e}_{i,k}-y_{i,k})-\frac{2}{N}(l_{i,k}-y_{i,k})\\
            &=\frac{2}{N}(l^{\rm e}_{i,k}-l_{i,k})\\
            &= -\frac{2 \epsilon c_{p,k}}{N} \qquad \qquad \rhd \text{Eq.~\eqref{defference_of_labels}} \\
            % & \\
            &= 
            \left\{
                \begin{aligned}
                    &-\frac{2 \epsilon c_{p,p}}{N} =\frac{2 \epsilon}{N}(1-s'_{p,p}) > 0, &&p=k\\
                    &-\frac{2 \epsilon c_{p,k}}{N} = -\frac{2 \epsilon}{N} s'_{p,k} < 0, &&p\neq k
                .\end{aligned}
            \right. 
    \end{aligned} 
    \label{DBP_MSE}
\end{equation}
So, there are two conclusions which are similar to Section \ref{situation_with_CE}:

\begin{enumerate}
\item 
Compared with one-hot label, enhancement label makes $y_{i,p}$ increase more and $y_{i,k}$ ($k\neq p$) decrease more.
\item 
Suppose there is an instance $a$ belonging to a minority class $u$ ($a \in {\bf D}_u$) and an instance $b$ belonging to a majority class $v$ ($b \in {\bf D}_v$), then:
\begin{gather}
    s'_{u,u}<s'_{v,v}\ 
,\\
    s'_{u,v}>s'_{v,u}\ 
.\end{gather}
And:
\begin{gather}
    \text{DBP}_{a,u}>\text{DBP}_{b,v}>0
,\\
    \text{DBP}_{a,v}<\text{DBP}_{b,u}<0
.\end{gather}
That means $y_{a,u}$ increases more than $y_{b,v}$ and $y_{a,v}$ decreases more than $y_{b,u}$.
\end{enumerate}

\section{The Imbalanced Setting of MNIST}
\label{imbalanced_setting_of_mnist}

\begin{figure}[h]
\centering
\vspace{-0.5cm}
\subfloat[10\% of Even Classes]
    {\centering
     \includegraphics[width=0.43\textwidth,trim = 0. 0. 0. 33, clip]{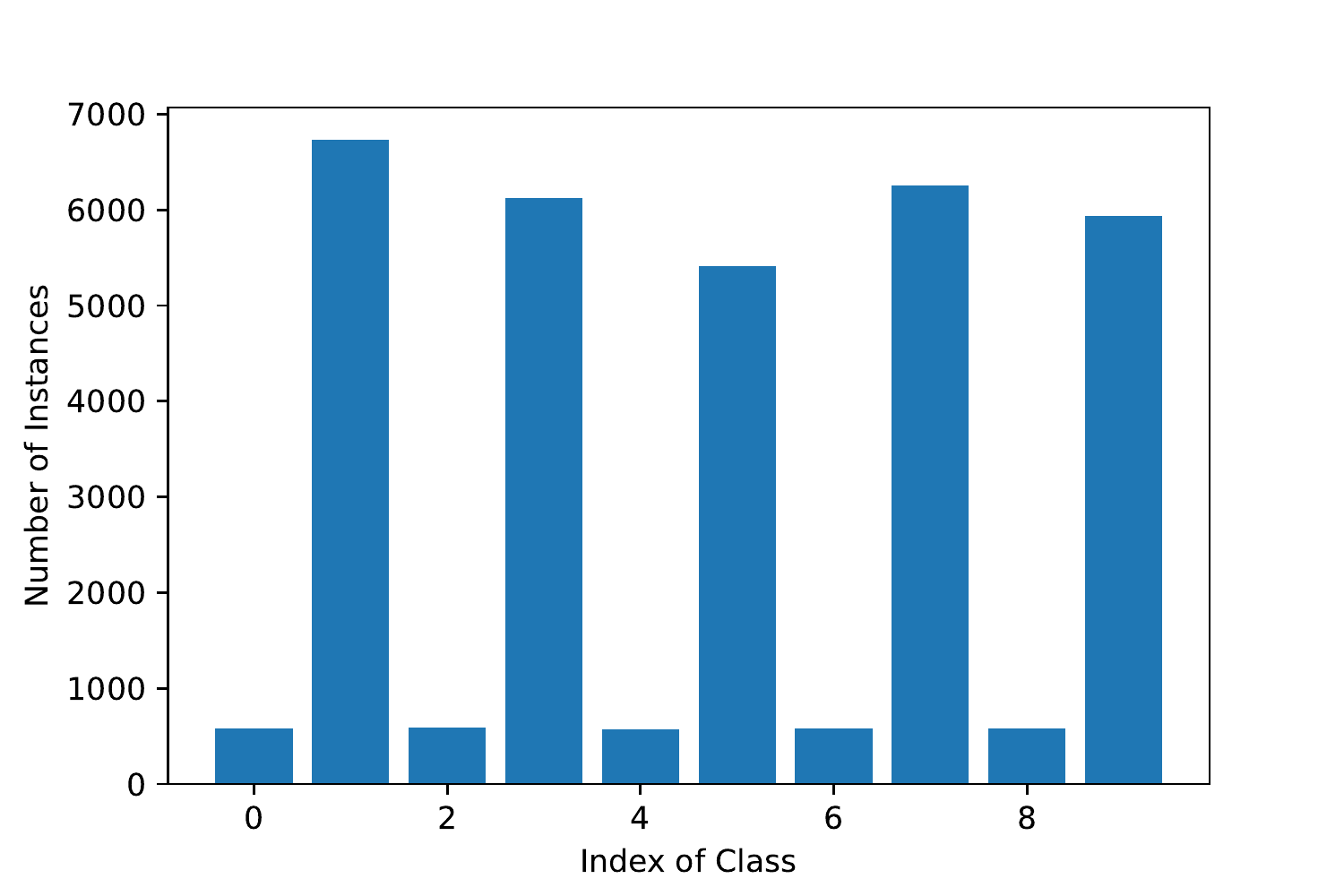}
    }
\qquad
\subfloat[Long-Tailed Distribution]
    {\centering
     \includegraphics[width=0.43\textwidth,trim = 0. 0. 0. 33, clip]{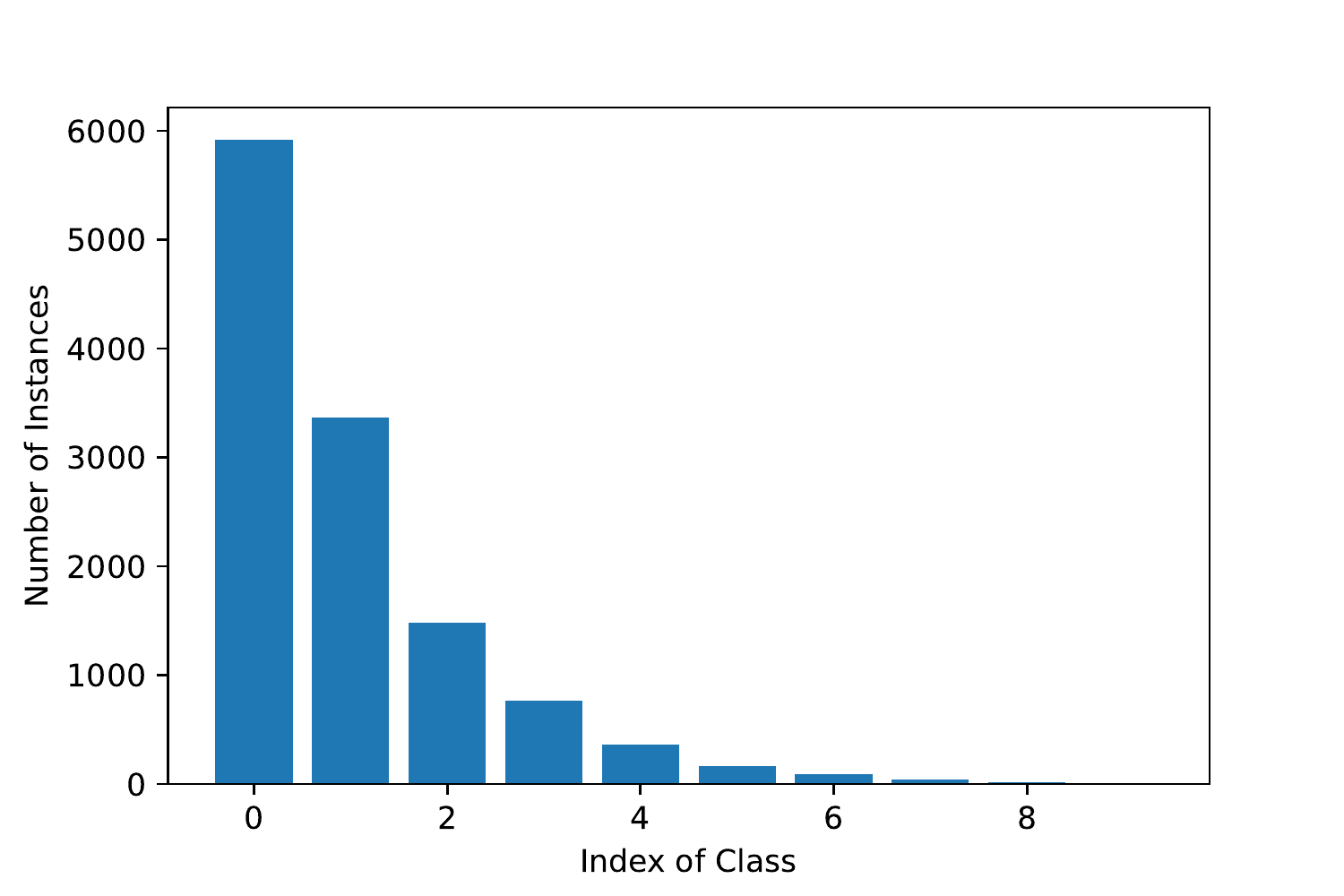}
    }
\caption{The training sets of imbalanced MNISTs. The original data set is balanced. To evaluate our algorithm, we manually unbalance the data.}
\label{MNIST_imbalance}
\end{figure}
The distributions of MNIST can be described as:
\begin{equation}
    n^{\text{tr}\, *}_p = \left\lceil \frac{n^{\text tr}_p}{f_p} \right\rceil - n^{\text v}_p 
\end{equation}
\noindent where $n^{\text tr}_p$ denotes the number of training images belonging to class $p$ before adjustment and $n^{\text{tr}\, *}_p$ denotes the number after adjustment, $n^{\text v}_p$ denotes the number of validation samples per class. We adjust the data set in two ways: 
\begin{align}
    &f_p = 
        \left\{
            \begin{aligned}
                & 9[(p+1) \text{ mod } 2]+1, &&\text{10\% of even classes} \\
                & 2^{\, p}, &&\text{long-tailed distribution} 
            \end{aligned}
        \right.\\
    &n^{\text v}_p = 
        \left\{
            \begin{aligned}
                & 10, &&\text{10\% of even classes} \\
                & 5, &&\text{long-tailed distribution} 
            \end{aligned}
        \right.
\end{align}
 where $p$ denotes the index of class and note that $0 \leq p \leq 9$. In the first one (10\% of even classes), we refer to the data distribution created by Khan et al. \cite{cosenCNN}. Because Khan's beginning index of class and ours are different, there are some statement differences. The distributions are shown in Figure \ref{MNIST_imbalance}. 
 
\section{Potential Negative Societal Impacts}

The imbalanced learning will help the minority groups get enough social attention, but it may also be used to identify them and cause discrimination (such as HIV patients). It's necessary to make sure imbalanced classification methods (including the enhancement encoding) will not harm the legal rights of minority groups.

\section{Computational Resources}

We train all models in this paper with the internal cluster and the GPU is NVIDIA Tesla V100S.

\end{document}